\documentclass[conference]{IEEEtran}
\IEEEoverridecommandlockouts
\usepackage[top=0.7in, bottom=1in, left=0.625in, right=0.625in]{geometry}
\usepackage{amsmath,amssymb,amsfonts}
\usepackage{graphicx}
\usepackage[style=numeric, sorting=none]{biblatex}
\usepackage{textcomp}
\usepackage{algorithm}
\usepackage{tablefootnote}
\usepackage{algpseudocode}
\usepackage[acronym]{glossaries}
\usepackage{xcolor}
\usepackage{placeins}
\usepackage{float}
\usepackage{microtype}
\usepackage{titlesec}
\usepackage{subcaption}
\usepackage{multirow}
\usepackage{booktabs}

\usepackage{pgf}
\usepackage{pgfmath}
\usepackage{arydshln}

\usepackage[
breaklinks=true,
colorlinks=true,
linkcolor=blue,
citecolor=blue,
urlcolor=blue,
filecolor=blue,
pdffitwindow,
bookmarks=true,
bookmarksopen=true,
bookmarksnumbered=true,
pdftitle={},
pdfauthor={},
pdfsubject={},
pdfkeywords={}
]{hyperref}

\usepackage{cleveref}
\crefformat{equation}{(#1)}
\crefformat{figure}{Figure\,#1}
\crefformat{section}{Section\,#1}
\crefformat{table}{Table\,#1}

\usepackage{changes}
\definecolor{YB}{RGB}{0,0,200}
\definecolor{EN}{RGB}{0,200,0}
\definechangesauthor[color=YB]{YB}
\definechangesauthor[color=EN]{EN}

\def\BibTeX{{\rm B\kern-.05em{\sc i\kern-.025em b}\kern-.08em
    T\kern-.1667em\lower.7ex\hbox{E}\kern-.125emX}}
\bibliography{bib/neuromophic.bib}
\titlespacing*{\section}{0pt}{1ex}{1ex}
\titlespacing*{\subsection}{0pt}{0.5ex}{0.5ex}

\begin{document}

\title{Sparse Axonal and Dendritic Delays Enable Competitive SNNs for Keyword Classification\\
\thanks{This project is funded by the German Federal Ministry of Research, Technology, and Space, NEUROTEC-II grant no.~16ME0398K, 16ME0399}
}

\author{
\IEEEauthorblockN{Younes Bouhadjar}
\IEEEauthorblockA{\textit{Forschungszentrum Jülich} \\
Jülich, Germany \\
y.bouhadjar@fz-juelich.de}
\and
\IEEEauthorblockN{Emre Neftci}
\IEEEauthorblockA{\textit{Forschungszentrum Jülich} \\
Jülich, Germany \\
\textit{RWTH Aachen, Germany}\\
e.neftci@fz-juelich.de}
}

\maketitle


\begin{abstract}
Training transmission delays in spiking neural networks (SNNs) has been shown to substantially improve their performance on complex temporal tasks. In this work, we show that learning either axonal or dendritic delays enables deep feedforward SNNs composed of leaky integrate-and-fire (LIF) neurons to reach accuracy comparable to existing synaptic delay learning approaches, while significantly reducing memory and computational overhead. SNN models with either axonal or dendritic delays achieve up to $95.58\%$ on the Google Speech Command (GSC) and $80.97\%$ on the Spiking Speech Command (SSC) datasets, matching or exceeding prior methods based on synaptic delays or more complex neuron models. By adjusting the delay parameters, we obtain improved performance for synaptic delay learning baselines, strengthening the comparison. We find that axonal delays offer the most favorable trade-off, combining lower buffering requirements with slightly higher accuracy than dendritic delays. We further show that the performance of axonal and dendritic delay models is largely preserved under strong delay sparsity, with as few as 20\% of delays remaining active, further reducing buffering requirements. Overall, our results indicate that learnable axonal and dendritic delays provide a resource-efficient and effective mechanism for temporal representation in SNNs. Code will be made available publicly upon acceptance.
Code is available at \url{https://github.com/YounesBouhadjar/AxDenSynDelaySNN}
\end{abstract}


\section{Introduction}
Theoretical studies have shown that transmission delays unlock a rich repertoire of spatiotemporal representations in spiking neural networks (SNNs).
Delays enable neurons to selectively respond to or replay precise spatiotemporal patterns \cite{Izhikevich06_polychronization} as well as expand the class of functions they can represent beyond those achievable by purely weight-based threshold circuits \cite{Maass99_temporal}.
Building on these theoretical insights, recent advances made it possible to train SNNs with delays to solve challenging temporal benchmarks \cite{Hammouamri24_dcls, Xu25_ARSC, Queant25_delrec, Mszros2025_delayprop, Goeltz25_delgrad}.
Models with learnable delays in feedforward or recurrent architectures \cite{Hammouamri24_dcls, Queant25_delrec, Mszros2025_delayprop} achieve performance comparable to SNNs, which rely instead on adaptive neuron dynamics \cite{Bittar22_adLIF_RadLIF, Deckers24_cAdLIF, Fabre25_SiLIF}. 
However, these results are primarily obtained using synaptic or recurrent delays, both of which incur substantial computational and memory overhead. 
In particular, synaptic delays require parameters and buffering at the level of individual synapses, increasing memory and buffering requirements, which can limit scalability and hinder deployment on neuromorphic hardware \cite{Wang18,Zapata20_axonal_delay, Khodamoradi21_s2n2,PatinoSaucedo23_axonal_delays, Zhao23_axonalonfpga, Patino-Saucedo24_synaptic_delays, Meijer25_synaptic_delays, Leone25_syntzulu_axonal_delay}.
\par
Axonal and dendritic delays provide a more favorable alternative in which delays are associated with neurons rather than individual synapses. 
In particular, axonal delays assign a single delay shared across all outgoing connections of a neuron, directly corresponding to axonal conduction delays observed in biological neurons \cite{Bucci25}. 
Dendritic delays similarly associate a single delay with all incoming connections of a neuron, modeling propagation along the dendritic tree \cite{Ol_h_2025}. 
Both mechanisms significantly reduce the number of delay parameters and buffering requirements compared to synaptic delays.
However, existing feedforward SNNs with axonal delays report lower accuracy than synaptic delay-based models \cite{Sun23_learned_axonal, Sun25_quantDelay}, and dendritic delays have so far been explored only in small-scale settings \cite{Goeltz25_delgrad}.
\par
In this work, we systematically study axonal and dendritic delays in deep feedforward SNNs.
We show that both mechanisms achieve accuracies close to those of synaptic delay learning models on keyword classification datasets, despite relying on more constrained formulations. 
We show further that axonal delays require fewer buffering resources and achieve slightly higher accuracy than dendritic delays across tasks. 
Our results substantially narrow the performance gap between axonal and synaptic delays and improve upon prior axonal delay-based models. 
Furthermore, we demonstrate that axonal and dendritic delays can be sparsified jointly with synaptic weights with only a minor impact on accuracy.
\par
Overall, our findings clarify the functional roles and trade-offs of different delay mechanisms in SNNs and show that axonal and dendritic delays enable effective temporal computation with reduced parameter count, memory usage, and buffering requirements, making them particularly well suited for efficient neuromorphic implementations.
\section{Methods}
The spiking neuron model follows leaky integrate-and-fire (LIF) dynamics. The membrane potential \(U_i(t)\) of neuron \(i\) evolves according to
\begin{equation}
    U_i(t + 1) = \beta\, U_i(t) + I_i(t),
\end{equation}
where \(0 < \beta < 1\) is the decay factor and \(I_i(t)\) is the synaptic input current. A spike is generated when the membrane potential reaches the threshold \(U_{\text{th}}\), i.e., \(S_i(t)=1\) if \(U_i(t)\ge U_{\text{th}}\) and \(S_i(t)=0\) otherwise. After a spike, the membrane potential is reset as
\begin{equation}
    U_i(t) = (1 - S_i(t))\, U_i(t).
\end{equation}
\par
We build on the convolution-based formulation of synaptic delays for feedforward SNNs proposed in \cite{Hammouamri24_dcls}.
Axonal delays are implemented by assigning a common delay kernel to all outgoing synapses of a presynaptic neuron $j$, while dendritic delays use a common kernel for all incoming synapses to a postsynaptic neuron $i$.
We denote the synaptic weight from presynaptic neuron $j$ to postsynaptic neuron $i$ by $w_{ij}$.
We assign each connection a delay $d_{ij}$; for axonal delays we tie delays across outgoing synapses so that $d_{ij}=d^{\mathrm{ax}}_{j}$, and for dendritic delays we tie delays across incoming synapses so that $d_{ij}=d^{\mathrm{den}}_{i}$.
For $H$ presynaptic neurons, the synaptic current to neuron $i$ follows
\begin{equation}
    I_i(t)=\sum_{j=1}^{H} w_{ij}\, S_j\!\bigl(t-d_{ij}\bigr)
    \;=\;\sum_{j=1}^{H} \bigl(k_{ij}\ast S_j\bigr)(t),
\end{equation}
where \(\ast\) denotes discrete-time convolution and \(k_{ij}\) is the associated delay kernel.
\par
Delays are learned using the one-dimensional DCLS formulation from \cite{Hassani23_dcls} with a Gaussian kernel centered at \(d_\text{max}-d_{ij}-1\) and standard deviation $\sigma$. Here, $d_\text{max}$ is the kernel size and is chosen as the maximum delay plus one.
The number of groups, corresponding to the channel grouping in the convolution operation, is set to the number of presynaptic neurons for axonal delays and to the number of postsynaptic neurons for dendritic delays.
\par
To restrict DCLS to only learning delays, we fix $w_{ij}=1$. Trainable synaptic weights are reintroduced with a linear layer placed after the DCLS module.
Following \cite{Hammouamri24_dcls}, all standard deviations $\sigma$ are initialized to $T_{\text d}/2$ at the start of training and are gradually annealed to $0.5$ over the initial phase of training (first 25\% of epochs).
At inference time, the learned delays are discretized by rounding to the nearest integer, again following \cite{Hammouamri24_dcls}.

\section{Results}

\subsection{Methodology}

The architecture stacks $L$ layers, each combining a DCLS layer, an MLP, batch normalization, and a LIF layer with $H$ neurons.
During training, spike non-differentiability is handled using the ATan surrogate gradient with slope parameter $a$.
The readout mechanism uses the membrane potential of a leaky integrator.
\Cref{tab:network_parameters} lists the key hyperparameters, with default values highlighted in bold and used throughout unless stated otherwise.
\par
We evaluate our models on keyword classification tasks using the Google Speech Commands v0.02 (GSC) dataset \cite{Warden18_GSC} and its event-based counterpart \cite{Cramer22_spiking_datasets}, the Spiking Speech Commands (SSC), following standard protocols in recent neuromorphic learning studies \cite{Bittar22_adLIF_RadLIF}.
SSC is obtained by converting GSC audio signals into streams of binary spike events using the cochlea model of \cite{Cramer22_spiking_datasets}. The resulting event streams are discretized using fixed 10 ms time bins and spatially binned from 700 into 140 input channels.
For GSC, audio waveforms are processed using 10 ms frames and passed through a Mel filterbank with 40 Mel filters. Notably, the original DCLS synaptic-delay formulation uses 140 Mel-frequency channels, which we adopt as a reference configuration for direct comparison.
\par
After conducting a hyperparameter search, summarized in \cref{tab:learning_parameters}, we observed that the best-performing configurations for both axonal and dendritic delays closely align with those reported in \cite{Hammouamri24_dcls}.
We use the Adam optimizer with separate parameter groups, setting the base learning rate to 0.001 for synaptic weights and 0.1 for delays.
For optimization, we apply a one-cycle learning rate scheduler to the synaptic weights and use cosine annealing for the delay learning rates.
\par
Weight sparsity and delay sparsity are controlled using fixed binary masks: 
$\kappa$ denotes the weight sparsity level and $\eta$ denotes the delay sparsity level.
These masks are kept fixed during training and evaluation.

\begin{table*}[h]
    \centering
    \caption{Network parameters for different datasets. Bold numbers depict the default value.}
    \begin{tabular}{l c c c c c c}
    \hline
    \textbf{Dataset} & \# Hidden layers ($L$) & \# Hidden size ($H$) & Decay coefficient ($\beta$) & Max delay ($d_\text{max}$) & Threshold ($U_\text{th}$) & Surrogate slope ($a$)\\
    \hline
    GSC/SSC & 3 & 512 & 0.33 & $\{5,11,\mathbf{15},31,35,41\}$ & 1 & 5\\
    \hline
    \end{tabular}
    \label{tab:network_parameters}
\end{table*}

\begin{table*}[h]
    \centering
    \caption{Learning parameters and hyperparameter search space for different datasets. Bold numbers depict the best hyperparameters.}
    \begin{tabular}{l c c c c c c c}
    \hline
    \textbf{Datasets} & \# Epochs & Batch size &
    LR (weights) &
    LR (delays) &
    Scheduler (weights) &
    Scheduler (delays) &
    Dropout \\
    \hline
    GSC/SSC & 100 & 512 &
    $\{10^{-2},\,\mathbf{10^{-3}},\,10^{-4}\}$ &
    $\{\mathbf{10^{-1}},\,10^{-2},\,10^{-3}\}$ &
    \{\textbf{one-cycle}, cosine, none\} &
    \{one-cycle, \textbf{cosine}, none\} &
    0.25 \\
    \hline
    \end{tabular}
    \label{tab:learning_parameters}
\end{table*}

\subsection{Metrics}
\label{sec:metrics}
We evaluate models using three main metrics: the total number of spikes, the number of synaptic operations (SOPs), and the buffer size. The number of spikes is the total count of spikes generated by all spiking layers. SOPs only account for spike events that propagate through synaptic weights and exclude operations related to the delay mechanism. This choice is motivated by two considerations: we assume a sufficiently simple delay mechanism, whose primary effect is captured by the buffer-size metric introduced below, and delay handling mainly involves inexpensive buffering and indexing of spike events that are already required in spiking neural networks without explicit delays.
\par
Implementing transmission delays in SNNs admits multiple buffering and scheduling strategies, depending on the delay mechanism and the underlying hardware architecture. Optimized implementations can reduce memory usage through shared routing structures or hardware-specific scheduling. In this work, we adopt a simple, non-optimized buffering strategy to estimate memory costs in a consistent and architecture-agnostic manner. This choice provides a transparent upper-bound estimate and allows fair comparison across different delay mechanisms without relying on hardware-specific assumptions.
\par
We measure the required buffer size to store the delayed activity as
\begin{equation}
    S_{\text{buffers}}
    = \sum_l H \cdot s
    + C \times d_{\text{max}},    
\end{equation}
where the first term accounts for storing neuron states and the second term accounts for buffering delayed spike-related information over a delay window of length $d_{\text{max}}$. Here, $s$ denotes the storage size of a neuron state, such as the membrane potential, and $C$ is a coefficient that depends on the delay mechanism and buffering strategy.
\par
We distinguish between unshared and shared buffering strategies. In the unshared-buffer case, buffering is performed at the neuron level. For synaptic and axonal delays, each presynaptic neuron maintains its own buffer to store emitted spikes until their associated delays are reached. For dendritic delays, buffering is performed at the postsynaptic neuron, where incoming spike weights are accumulated in a neuron-level ring buffer.
In the shared-buffer case, delayed events from all neurons in a layer are stored in a common buffer shared across the layer. Each buffer entry must therefore additionally encode the identity of the source neuron for synaptic delays or axonal delays, or the identity of the target neuron for dendritic delays.
\par
Under these assumptions, the coefficient $C$ takes different forms depending on the delay mechanism and buffering strategy, as summarized in \cref{tab:C_coefficients}.
As indicated in this latter table, estimating buffer sizes requires introducing variables to account for the storage size of the weight $v$ and the address $m$.
We use a one-state Leaky Integrate-and-Fire neuron, storing only the membrane potential, with state size $s = 16$ bits. Accumulated synaptic values are stored using the same precision, giving $v = 16$ bits. For shared buffers, the address size is defined as $m = \lceil \log_2(H) \rceil$. These assumptions provide a consistent and hardware-oriented basis for comparing memory costs across delay mechanisms.

\begin{table}[t]
    \centering
    \caption{Effective buffering coefficient $C$ for different delay mechanisms and buffering strategies.
    $H$ is the number of neurons within a layer, $v$ the storage size of a synaptic weight value, and $m$ the address size.
    The variables $\rho_\text{n}$ and $rho_\text{p}$ represent the spike rates at the single-neuron and population levels, respectively, computed as the number of generated spikes divided by the maximum delay duration $d_\text{max}$.
    These rates determine the overall buffer occupancy. Lower spike rates imply lower occupancy and therefore permit smaller buffer sizes.
    Since neurons in the networks considered in this project tend to emit spikes in short bursts, we conservatively set $\rho_\text{n} = 1$.
    For shared buffers, we use $\rho_\text{p} = 0.2$, as observed empirically in our experiments.}
    \label{tab:C_coefficients}
    \begin{tabular}{lcc}
    \toprule
    Delay mechanism & Unshared buffering & Shared buffering \\
    \midrule
    Synaptic
    & $H^{2}\,\rho_\text{n}$
    & $m\,H^{2}\,\rho_\text{p}$ \\
    Axonal
    & $H\,\rho_\text{n}$
    & $m\,H\,\rho_\text{p}$ \\
    Dendritic
    & $v\,H\,\rho_\text{n}$
    & $(v + m)\,H\,\rho_\text{p}$ \\
    \bottomrule
    \end{tabular}
\end{table}


\subsection{Benchmarking results}

In this section, we report the performance and computational cost of our models on the SSC and GSC benchmarks (see \cref{tab:performance_models_comparison}). Across both datasets, SNN models with learnable axonal or dendritic delays achieve accuracy comparable to synaptic-delay models and, in several configurations, slightly exceed adaptive-neuron model baselines.
\par
\begin{figure}
    \centering
    \includegraphics[width=\linewidth]{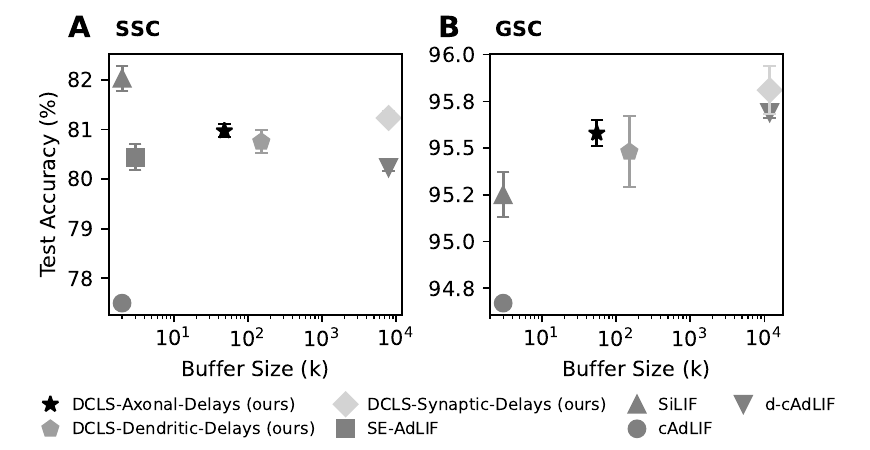}
    \caption{Test accuracy as a function of buffer size for (\textbf{A}) SSC and (\textbf{B}) GSC benchmarks. Results compare DCLS-based models with axonal, dendritic, and synaptic delays (ours) against baseline spiking neuron models, including SiLIF, cAdLIF, d-cAdLIF, and SE-AdLIF.
    The selected DCLS-based models correspond to the best-performing configurations reported in \cref{tab:performance_models_comparison}.
    Buffer size is shown on a logarithmic scale. Points denote mean test accuracy across runs, with error bars indicating standard deviation.}
    \label{fig:test_acc_vs_buffer}
\end{figure}

On SSC, axonal-delay models reach up to $80.97\%$ accuracy, closely matching synaptic delay learning while using roughly half the number of parameters and orders of magnitude smaller buffer sizes (see \cref{fig:test_acc_vs_buffer}A). Dendritic-delay models achieve similar accuracy, but incur larger buffer sizes than axonal-delay models due to neuron-level accumulation of delayed weights (see \cref{sec:metrics}).
\par
Compared to adaptive LIF variants such as cAdLIF, SE-adLIF, and SiLIF, axonal and dendritic delay models achieve comparable accuracy while relying on standard LIF dynamics and fewer neuron-level state variables. In contrast, models such as SiLIF and SE-adLIF reach strong accuracy at the cost of increased numbers of simulation steps, synaptic operations, or spikes, whereas delay-based models, especially axonal-delay variants, maintain competitive performance with moderate buffer sizes and similar orders of magnitude of synaptic operations.
DelRec achieves state-of-the-art accuracy; however, its reliance on recurrent delays may introduce additional buffering requirements, increase synaptic operations, and complicate neuromorphic implementation compared to feedforward SNN architectures.
\par
On GSC, axonal and dendritic delay models consistently achieve accuracies above $95\%$, comparable to both synaptic delay learning and adaptive neuron baselines. While synaptic delay models require very large buffer sizes, axonal and dendritic delay models reduce buffering to the order of thousands (see \cref{fig:test_acc_vs_buffer}B). Among neuron-level delay mechanisms, axonal-delay models require smaller buffers and generate fewer spikes than dendritic-delay variants, suggesting a more favorable fit for neuromorphic hardware implementations \cite{Yik25_neurobench}.
\par
\begin{figure}
    \centering
    \includegraphics[width=\linewidth]{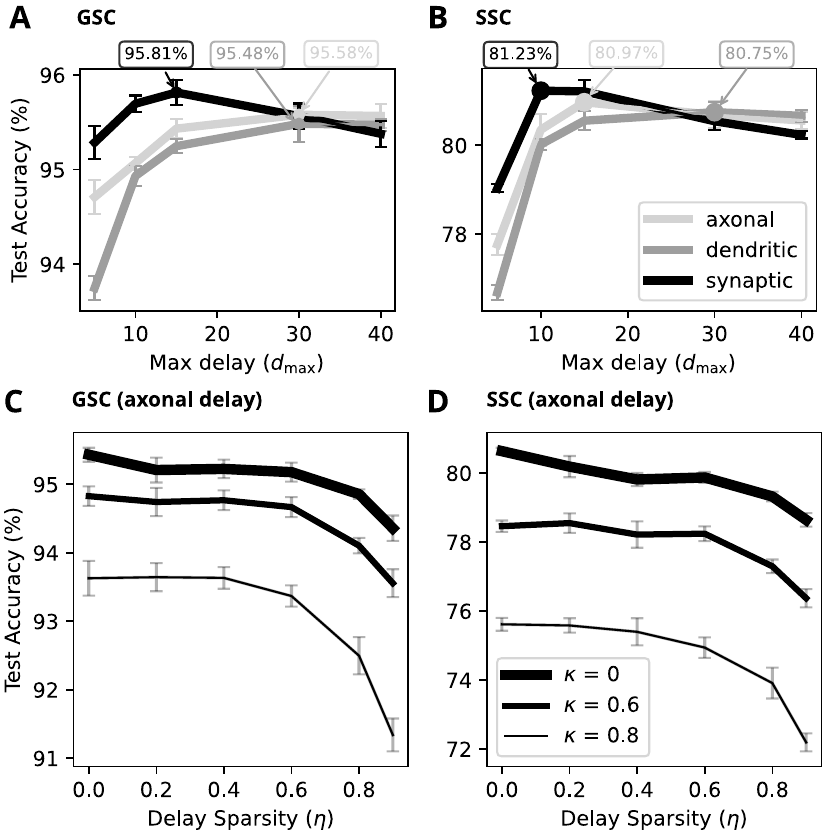}
    \caption{Effect of delay range and sparsity on classification accuracy. \textbf{(A--B)} Test accuracy as a function of the maximum allowed delay $d_{\max}$ for synaptic, axonal, and dendritic delay models, evaluated on GSC (A) and SSC (B).
    \textbf{(C--D)} Test accuracy of axonal-delay models as a function of delay sparsity $\eta$ for different levels of weight sparsity $\kappa$, evaluated on GSC (C) and SSC (D).}
    \label{fig:test_acc_delay_analysis}
\end{figure}

We note that the synaptic delay learning baselines reported in \cref{tab:performance_models_comparison} achieve higher accuracy than previously reported results \cite{Hammouamri24_dcls}. This improvement is due to a more careful tuning of the maximum delay range $d_{\max}$, whose effect on performance is demonstrated in \cref{fig:test_acc_delay_analysis}.
In addition, for the GSC benchmark, reducing the dimensionality of the Mel spectrogram input from 140 to 40 frequency bins yields a small performance gain and a decrease in synaptic operations. These refinements strengthen the synaptic delay baselines and enable a fairer comparison with axonal and dendritic delay models.
\par
We observe that shared and unshared buffering strategies (see \cref{sec:metrics}) lead to similar memory costs for axonal and synaptic delays. In the shared-buffer case, buffer slots are reused across neurons, reducing the number of required slots in proportion to the effective spike rate and delay sparsity. However, this gain is largely compensated for by the need to store additional address information for each buffered event. Under the sparse spiking regimes considered in this work, these two effects balance out, resulting in comparable overall memory costs for shared and unshared buffers. In contrast, for dendritic delays, the unshared buffering strategy is significantly more costly, as it requires neuron-level accumulation of weighted inputs in dedicated ring buffers. This accumulation increases the stored data per buffer slot and cannot be offset by buffer reuse, leading to substantially higher memory usage compared to the shared-buffer implementation.
\par
We further observe that strong sparsification of axonal delays has only a moderate impact on performance. Even with $80\%$ delay sparsity and additional weight sparsity, accuracy is only slightly degraded on both SSC and GSC, while buffering requirements are further reduced.
Introducing weight sparsity leads to a further decrease in accuracy, which nevertheless remains within a reasonable range.
This degradation is accompanied by a substantial reduction in synaptic operations, highlighting an attractive accuracy–efficiency trade-off for resource-constrained settings.
Increasing network width to match the parameter count partially recovers the performance, leaving only a small residual gap relative to delay-only sparsification.


\pgfmathsetmacro{\asize}{5}
\pgfmathsetmacro{\wsize}{16}
\pgfmathsetmacro{\ssize}{16}

\pgfmathsetmacro{\DSut}{1} 
\pgfmathsetmacro{\DSt}{0.2} 
\pgfmathsetmacro{\DSp}{0.8} 



\pgfmathsetmacro{\DSSCsyn}{10}
\pgfmathsetmacro{\DSSC}{15}
\pgfmathsetmacro{\DSSCb}{30}
\pgfmathsetmacro{\LSSC}{3}
\pgfmathsetmacro{\NhSSC}{512 / 1000}
\pgfmathsetmacro{\NhSSCse}{720 / 1000}

\pgfmathsetmacro{\NbSSCax}{round(\LSSC * \NhSSC * (\ssize + \asize * \DSSC * \DSt))}
\pgfmathsetmacro{\NbSSCaxb}{round(\LSSC * \NhSSC * (\ssize + \asize * \DSSCb * \DSt))}
\pgfmathtruncatemacro{\NbSSCax}{\NbSSCax}
\pgfmathtruncatemacro{\NbSSCaxb}{\NbSSCaxb}

\pgfmathtruncatemacro{\NbSSCaxS}{\NbSSCax * \DSp}
\pgfmathtruncatemacro{\NbSSCaxSb}{\NbSSCax * \DSp * 1.62}

\pgfmathsetmacro{\NbSSCaxu}{round(\LSSC * \NhSSC * (\ssize + \DSSC * \DSut))}
\pgfmathsetmacro{\NbSSCaxub}{round(\LSSC * \NhSSC * (\ssize + \DSSCb * \DSut))}
\pgfmathtruncatemacro{\NbSSCaxu}{\NbSSCaxu}
\pgfmathtruncatemacro{\NbSSCaxub}{\NbSSCaxub}

\pgfmathtruncatemacro{\NbSSCaxuS}{\NbSSCaxu * \DSp}
\pgfmathtruncatemacro{\NbSSCaxuSb}{\NbSSCaxu * \DSp * 1.62}


\pgfmathtruncatemacro{\LSSCtwostate}{2}
\pgfmathsetmacro{\NbSSCxadLIF}{round(\LSSCtwostate * \NhSSC * 2)}
\pgfmathsetmacro{\NbSSCseadLIF}{round(\LSSCtwostate * \NhSSCse * 2)}
\pgfmathtruncatemacro{\NbSSCxadLIF}{\NbSSCxadLIF}
\pgfmathtruncatemacro{\NbSSCseadLIF}{\NbSSCseadLIF}


\pgfmathsetmacro{\NbSSCdcadlif}{round((\LSSCtwostate * \NhSSC * 2 / 1000 * \ssize + \LSSCtwostate * \NhSSC * \NhSSC * \DSSC * \asize * \DSt)}
\pgfmathtruncatemacro{\NbSSCdcadlif}{\NbSSCdcadlif}   

\pgfmathsetmacro{\NbSSCdcadlifu}{round(\LSSCtwostate * \NhSSC * 2 * \ssize  / 1000 + \LSSCtwostate * \NhSSC * \NhSSC * \DSSC)}             
\pgfmathtruncatemacro{\NbSSCdcadlifu}{\NbSSCdcadlifu}



\pgfmathsetmacro{\NbSSCsyn}{round((\LSSC * \NhSSC * \NhSSC * \DSSCsyn * \asize * \DSt))}
\pgfmathsetmacro{\NbSSCsynb}{round((\LSSC * \NhSSC * \NhSSC * \DSSCb * \asize * \DSt))}
\pgfmathsetmacro{\NbSSCsynbb}{round((\LSSC * \NhSSC * \NhSSC * 25 * \asize * \DSt))}
\pgfmathtruncatemacro{\NbSSCsyn}{\NbSSCsyn}
\pgfmathtruncatemacro{\NbSSCsynb}{\NbSSCsynb}
\pgfmathtruncatemacro{\NbSSCsynbb}{\NbSSCsynbb}

\pgfmathsetmacro{\NbSSCsynu}{round((\LSSC * \NhSSC * \NhSSC * \DSSCsyn))}
\pgfmathsetmacro{\NbSSCsynub}{round((\LSSC * \NhSSC * \NhSSC * \DSSCb))}
\pgfmathtruncatemacro{\NbSSCsynu}{\NbSSCsynu}
\pgfmathtruncatemacro{\NbSSCsynub}{\NbSSCsynub}


\pgfmathsetmacro{\NbSSCden}{round(\LSSC * \NhSSC * \ssize + \DSSC * \DSt * (\wsize + \asize))}
\pgfmathsetmacro{\NbSSCdenb}{round(\LSSC * \NhSSC * \ssize + \DSSCb * \DSt * (\wsize + \asize))}
\pgfmathtruncatemacro{\NbSSCden}{\NbSSCden}
\pgfmathtruncatemacro{\NbSSCdenb}{\NbSSCdenb}

\pgfmathtruncatemacro{\NbSSCdenS}{\NbSSCden * \DSp}

\pgfmathsetmacro{\NbSSCdenu}{round(\LSSC * \NhSSC * \ssize + \DSSC * \DSut * \wsize)}
\pgfmathsetmacro{\NbSSCdenub}{round(\LSSC * \NhSSC * \ssize + \DSSCb * \DSut * \wsize)}
\pgfmathtruncatemacro{\NbSSCdenu}{\NbSSCdenu}
\pgfmathtruncatemacro{\NbSSCdenub}{\NbSSCdenub}

\pgfmathtruncatemacro{\NbSSCdenuS}{\NbSSCdenu * \DSp}



\pgfmathsetmacro{\DGSCsyn}{15}
\pgfmathsetmacro{\DGSC}{15}
\pgfmathsetmacro{\DGSCb}{30}
\pgfmathsetmacro{\LGSC}{3}
\pgfmathsetmacro{\NhGSC}{512 / 1000}

\pgfmathsetmacro{\NbGSCax}{round(\LGSC * \NhGSC * \ssize + \asize * \DGSC * \DSt)}
\pgfmathsetmacro{\NbGSCaxb}{round(\LGSC * \NhGSC * \ssize + \asize * \DGSCb * \DSt)}
\pgfmathtruncatemacro{\NbGSCax}{\NbGSCax}
\pgfmathtruncatemacro{\NbGSCaxb}{\NbGSCaxb}

\pgfmathtruncatemacro{\NbGSCaxS}{\NbGSCax * \DSp}
\pgfmathtruncatemacro{\NbGSCaxSb}{\NbGSCax * \DSp * 1.59}

\pgfmathsetmacro{\NbGSCaxu}{round(\LGSC * \NhGSC * \ssize + \DGSC * \DSut)}
\pgfmathsetmacro{\NbGSCaxub}{round(\LGSC * \NhGSC * \ssize + \DGSCb * \DSut)}
\pgfmathtruncatemacro{\NbGSCaxu}{\NbGSCaxu}
\pgfmathtruncatemacro{\NbGSCaxub}{\NbGSCaxub}

\pgfmathtruncatemacro{\NbGSCaxuS}{\NbGSCaxu * \DSp}
\pgfmathtruncatemacro{\NbGSCaxuSb}{\NbGSCaxu * \DSp * 1.59}


\pgfmathsetmacro{\NbGSCxadLIF}{round(\LGSC * \NhGSC * 2)}
\pgfmathtruncatemacro{\NbGSCxadLIF}{\NbGSCxadLIF}


\pgfmathsetmacro{\NbGSCdcadlif}{round(\LGSC * \NhGSC * 2 * \ssize / 1000 + \LGSC * \NhGSC * \NhGSC * \DGSC * \asize * \DSt)}
\pgfmathtruncatemacro{\NbGSCdcadlif}{\NbGSCdcadlif} 

\pgfmathsetmacro{\NbGSCdcadlifu}{round(\LGSC * \NhGSC * 2 * \ssize / 1000 + \LGSC * \NhGSC * \NhGSC * \DGSC)}
\pgfmathtruncatemacro{\NbGSCdcadlifu}{\NbGSCdcadlifu} 



\pgfmathsetmacro{\NbGSCsyn}{round((\LGSC * \NhGSC * \NhGSC * \DGSCsyn * \asize * \DSt))}
\pgfmathsetmacro{\NbGSCsynb}{round((\LGSC * \NhGSC * \NhGSC * \DGSCb * \asize * \DSt))}
\pgfmathsetmacro{\NbGSCsynbb}{round((\LGSC * \NhGSC * \NhGSC * 25 * \asize * \DSt))}
\pgfmathtruncatemacro{\NbGSCsyn}{\NbGSCsyn}
\pgfmathtruncatemacro{\NbGSCsynb}{\NbGSCsynb}
\pgfmathtruncatemacro{\NbGSCsynbb}{\NbGSCsynbb}

\pgfmathsetmacro{\NbGSCsynu}{round((\LGSC * \NhGSC * \NhGSC * \DGSCsyn))}
\pgfmathsetmacro{\NbGSCsynub}{round((\LGSC * \NhGSC * \NhGSC * \DGSCb))}
\pgfmathtruncatemacro{\NbGSCsynu}{\NbGSCsynu}
\pgfmathtruncatemacro{\NbGSCsynub}{\NbGSCsynub}


\pgfmathsetmacro{\NbGSCden}{round(\LGSC * \NhGSC * \ssize + \DGSC * \DSt * (\wsize + \asize))}
\pgfmathsetmacro{\NbGSCdenb}{round(\LGSC * \NhGSC * \ssize + \DGSCb * \DSt * (\wsize + \asize))}
\pgfmathtruncatemacro{\NbGSCden}{\NbGSCden}
\pgfmathtruncatemacro{\NbGSCdenb}{\NbGSCdenb}

\pgfmathtruncatemacro{\NbGSCdenS}{\NbGSCden * \DSp}

\pgfmathsetmacro{\NbGSCdenu}{round(\LGSC * \NhGSC * \ssize + \DGSC * \DSut * \wsize)}
\pgfmathsetmacro{\NbGSCdenub}{round(\LGSC * \NhGSC * \ssize + \DGSCb * \DSut * \wsize)}
\pgfmathtruncatemacro{\NbGSCdenu}{\NbGSCdenu}
\pgfmathtruncatemacro{\NbGSCdenub}{\NbGSCdenub}

\pgfmathtruncatemacro{\NbGSCdenuS}{\NbGSCdenu * \DSp}


\begin{table*}[h!]
\centering
\caption{Performance comparison of SNN models on the SSC and GSC datasets. Accuracies are reported as mean $\pm$ standard deviation over five runs. In addition, for each model we report the number of time steps per sequence sample, the number of model parameters, the total number of spikes generated by the network and the associated number of synaptic operations (SOPs) when processing a single sample, as well as the required buffer size (see \cref{sec:metrics}). The buffer size is given in bits, with values in parentheses corresponding to a shared buffer implementation and values outside parentheses to per-neuron (unshared) buffers. The parameters $d_\text{max}$, $\eta$, and $\kappa$ denote the maximum delay, delay sparsity, and weight sparsity, respectively.  * denotes our improved DCLS synaptic-delay baseline compared to the original work in \cite{Hammouamri24_dcls}.}
\begin{tabular}{lccccccc}
\hline
\textbf{Dataset} & \textbf{Method}                                & \textbf{Nb. steps}   & \textbf{\# Params} & \textbf{\# Buffer size}                   & \textbf{\# SOP} & \textbf{\# Spikes} & \textbf{Top1 Acc.} \\
\hline
\multirow{8}{*}{SSC}
 & cAdLIF   \cite{Deckers24_cAdLIF}                               & 100                   & 0.35M              & $\NbSSCxadLIF$k                          & 5.42M           & 3.99k              & 77.50\%                  \\
 & SE-AdLIF \cite{Baronig25_SEAdLIF}                              & 300                   & 1.6M               & $\NbSSCseadLIF$k                         & -               & -                  & 80.44 $\pm$ 0.26\%       \\
 & SiLIF    \cite{Fabre25_SiLIF}                                  & 300                   & 0.35M              & $\NbSSCxadLIF$k                          & 8.8M            & 20.43k             & 82.03 $\pm$ 0.25\%       \\
 & DelRec \cite{Queant25_delrec}                                  & 250                   & 0.37M              & -                                        & -               & -                  & 82.58 $\pm$ 0.08\%       \\ 
 & d-cAdLIF (synaptic delays) \cite{Deckers24_cAdLIF}             & 100                   & 0.7M               & \textbf{$\NbSSCdcadlif$M ($\NbSSCdcadlifu$M)} & 5.79M      & 5.32k              & 80.23 $\pm$ 0.07\%   \\
 & DCLS-Synaptic-Delays \cite{Hammouamri24_dcls}                  & 100                   & 1.2M               & \textbf{$\NbSSCsynb$M ($\NbSSCsynub$M)}   & 6.93M          & 11.89k             & 80.44 $\pm$ 0.11\%       \\ 
 \cdashline{2-8}
 & \textbf{DCLS-Synaptic-Delays* ($d_\text{max}=11$) ours}       & \textbf{100}           & \textbf{1.2M}      & \textbf{$\NbSSCsyn$M ($\NbSSCsynu$M)}     & \textbf{6.54M}  & \textbf{10.97k}    & \textbf{81.23 $\pm$ 0.13\%}    \\
 & \textbf{DCLS-Dendritic-Delays ($d_\text{max}=15$) ours}       & \textbf{100}           & \textbf{0.61M}     & \textbf{$\NbSSCden$k ($\NbSSCdenu$k)}     & \textbf{6.55M}  & \textbf{10.82k}    & \textbf{80.55 $\pm$ 0.21\%}    \\
 & \textbf{DCLS-Dendritic-Delays ($d_\text{max}=31$) ours}       & \textbf{100}           & \textbf{0.61M}     & \textbf{$\NbSSCdenb$k ($\NbSSCdenub$k)}   & \textbf{6.57M}  & \textbf{10.82k}    & \textbf{80.75 $\pm$ 0.23\%}    \\
 & \textbf{DCLS-Axonal-Delays ($d_\text{max}=15$) ours}          & \textbf{100}           & \textbf{0.61M}     & \textbf{$\NbSSCax$k  ($\NbSSCaxu$k)}      & \textbf{6.26M}  & \textbf{10.21k}    & \textbf{80.97 $\pm$ 0.13\%}    \\
 & \textbf{DCLS-Axonal-Delays ($d_\text{max}=31$) ours}          & \textbf{100}           & \textbf{0.61M}     & \textbf{$\NbSSCaxb$k  ($\NbSSCaxub$k)}    & \textbf{6.57M}  & \textbf{10.79k}    & \textbf{80.69 $\pm$ 0.31\%}    \\
 \cdashline{2-8}
 & \textbf{DCLS-Axonal-Delays ($\eta=0.8$) ours}                 & \textbf{100}           & \textbf{0.61M}     & \textbf{$\NbSSCaxS$k ($\NbSSCaxuS$k) }    & \textbf{6.36M}  & \textbf{10.53k}    & \textbf{79.32 $\pm$ 0.14\%}    \\
 & \textbf{DCLS-Axonal-Delays ($\eta=0.8$, $\kappa=0.6$) ours}   & \textbf{100}           & \textbf{0.36M}     & \textbf{$\NbSSCaxS$k ($\NbSSCaxuS$k) }    & \textbf{3.6M}   & \textbf{10.44k}    & \textbf{77.30 $\pm$ 0.2\%}     \\ 
 & \textbf{DCLS-Axonal-Delays ($\eta=0.8$, $\kappa=0.6$, $H=830$) ours}  & \textbf{100}   & \textbf{0.61M}     & \textbf{$\NbSSCaxSb$k ($\NbSSCaxuSb$k) }  & \textbf{9.18M} & \textbf{17.43k}    & \textbf{78.66 $\pm$ 0.28\%}     \\ 
\hline
\multirow{8}{*}{GSC}
 & cAdLIF   \cite{Deckers24_cAdLIF}                                & 100                  & 0.3M               & $\NbGSCxadLIF$k                           & 3.29M           & 5.12k              & 94.67\%                               \\
 & SiLIF    \cite{Fabre25_SiLIF}                                   & 98                   & 0.3M               & $\NbGSCxadLIF$k                           & 4M              & 6.6k               & 95.25\% $\pm$ 0.12\%                  \\
 & d-cAdLIF (synaptic delays) \cite{Deckers24_cAdLIF}              & 100                  & 0.61M              & $\NbGSCdcadlif$M ($\NbGSCdcadlifu$M)      & 3.77M           & 6.86k              & 95.69 $\pm$ 0.03                    \\
 & DCLS-Synaptic-Delays \cite{Hammouamri24_dcls}                   & 100                  & 1.2M               & $\NbGSCsynb$M ($\NbGSCsynub$M)            & 13M             & 12k                & 95.37\% $\pm$ 0.12\%              \\
\cdashline{2-8}
 & \textbf{DCLS-Synaptic-Delays* ($d_\text{max}=15$) ours}         & \textbf{100}         & \textbf{1.13M}     & \textbf{$\NbGSCsyn$M ($\NbGSCsynu$M)}     & \textbf{7.77M}  & \textbf{12.12k}    & \textbf{95.81\% $\pm$ 0.13\%}       \\
 & \textbf{DCLS-Dendritic-Delays  ($d_\text{max}=15$) ours}        & \textbf{100}         & \textbf{0.56M}     & \textbf{$\NbGSCden$k ($\NbGSCdenu$k)}     & \textbf{9.3M}   & \textbf{15.15k}    & \textbf{95.25\% $\pm$ 0.08\%}       \\
 & \textbf{DCLS-Dendritic-Delays ($d_\text{max}=31$) ours}         & \textbf{100}         & \textbf{0.56M}     & \textbf{$\NbGSCdenb$k ($\NbGSCdenub$k)}   & \textbf{10.05M} & \textbf{16.62k}    & \textbf{95.48\% $\pm$ 0.19\%}       \\
 & \textbf{DCLS-Axonal-Delays    ($d_\text{max}=15$) ours}         & \textbf{100}         & \textbf{0.56M}     & \textbf{$\NbGSCax$k ($\NbGSCaxu$k)}       & \textbf{8.1M}   & \textbf{12.69k}    & \textbf{95.43\% $\pm$ 0.1\%}        \\
 & \textbf{DCLS-Axonal-Delays   ($d_\text{max}=31$) ours}          & \textbf{100}         & \textbf{0.56M}     & \textbf{$\NbGSCaxb$k  ($\NbGSCaxub$k)}    & \textbf{8.78M}  & \textbf{13.97k}    & \textbf{95.58 $\pm$ 0.07\%}         \\
  \cdashline{2-8}
  & \textbf{DCLS-Axonal-Delays ($\eta=0.8$) ours}                  & \textbf{100}         & \textbf{0.56M}     & \textbf{$\NbGSCaxS$k ($\NbGSCaxuS$k)}     & \textbf{8.82M}  & \textbf{14.09k}    & \textbf{94.86 $\pm$ 0.07\%}         \\
 & \textbf{DCLS-Axonal-Delays ($\eta=0.8$, $\kappa=0.6$) ours}     & \textbf{100}         & \textbf{0.33M}     & \textbf{$\NbGSCaxS$k ($\NbGSCaxuS$k)}     & \textbf{5.34M}  & \textbf{14.98k}    & \textbf{94.10 $\pm$ 0.11\%}         \\ 
 & \textbf{DCLS-Axonal-Delays ($\eta=0.8$, $\kappa=0.6$, $H=816$) ours}  & \textbf{100}   & \textbf{0.56M}     & \textbf{$\NbGSCaxSb$k ($\NbGSCaxuSb$k)}   & \textbf{11.1M} & \textbf{20.49k}    & \textbf{94.7 $\pm$ 0.15\%}         \\ 
\hline
\end{tabular}
\label{tab:performance_models_comparison}
\end{table*}

\textbf{Effect of delay range and sparsity on performance.} For all delay types, increasing the delay range from small values ($d_\text{max}=5$) leads to a clear improvement in accuracy, indicating that access to a minimal temporal context is necessary to capture sufficient timing information (see \cref{fig:test_acc_delay_analysis}A,B). Performance typically peaks at moderate delay ranges and then saturates or slightly degrades as the delay range increases further. The reasons underlying this behavior remain unclear (see \cref{sec:limitations}). Providing the model with more temporal context than required by the task may lead to diminishing returns, as the additional capacity is not effectively exploited and may hinder optimization. Increased parameter counts and suboptimal hyperparameter tuning at larger delay ranges may further contribute to the observed performance saturation.
\par
Synaptic delays reach their optimal performance at slightly smaller $d_\text{max}$ compared to other delay models, consistent with their finer-grained, synapse-specific temporal resolution. In contrast, axonal and dendritic delays achieve comparable peak accuracy using neuron-level delays, suggesting that these more constrained mechanisms are sufficient to capture the timing structure required by the tasks when the delay range is appropriately chosen.
\par
Delay sparsity has a gradual and predictable effect on performance (see \cref{fig:test_acc_delay_analysis}C,D). As the fraction of active delays decreases, accuracy degrades slowly, demonstrating that only a subset of learned delays is required to preserve effective temporal representations. Even at high sparsity levels ($80\%$), the network retains a substantial fraction of its peak accuracy.
\par
Introducing weight sparsity leads to a further reduction in accuracy. Performance decreases consistently as both delay and weight sparsity increase, while a substantial reduction in synaptic operations is observed. Despite this degradation, performance remains within a reasonable range, indicating that temporal information conveyed by axonal delays can partially compensate for reduced synaptic connectivity. Overall, these results show that moderate delay ranges and sparse delay configurations are sufficient for accurate temporal processing, reinforcing the efficiency benefits of neuron-level delay mechanisms.

\begin{figure}
    \centering
    \includegraphics[width=\linewidth]{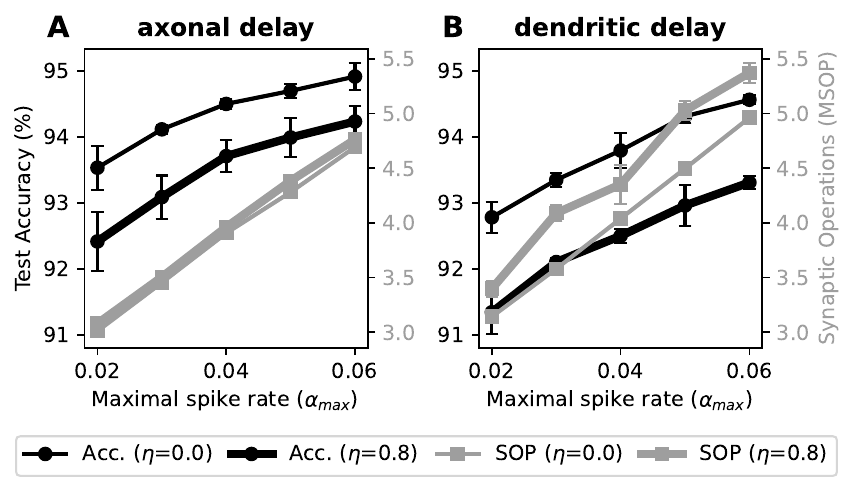}
    \caption{Effect of firing-rate regularization on axonal and dendritic delay models.
    Test accuracy (black) and synaptic operations (SOP, gray) as a function of the maximum allowed firing rate ($\alpha_\text{max}$) under activity regularization. The minimum firing rate ($\alpha_\text{min}$) is set to 0.001 and the regularization factor ($r$) to 0.5.
    \textbf{(A)} Axonal delays.
    \textbf{(B)} Dendritic delays.
    }
    \label{fig:test_acc_reg}
\end{figure}

\textbf{Effect of firing-rate regularization.} To restrict the firing rate of neurons, we use a firing-rate regularization that penalizes deviations from a target activity range \cite{Zenke21_sgl}. Let $f_\text{r}$ denotes the average firing rate of a neuron, defined as the total number of spikes emitted per neuron over the duration of an input sequence. The regularization is defined as
\begin{equation}
    \begin{aligned}
        R_{\text{quiet}} &= \sum_{i=1}^L \mathbf{1}^\top \mathrm{ReLU}\!\left(\alpha_{\min}\mathbf{1} - \mathbf{f}_r^{(i)}\right), \\
        R_{\text{burst}} &= \sum_{i=1}^L \mathbf{1}^\top \mathrm{ReLU}\!\left(\mathbf{f}_r^{(i)} - \alpha_{\max}\mathbf{1}\right), \\
        \mathcal{L}_{\text{reg}} &= r \left(R_{\text{quiet}} + R_{\text{burst}}\right).
    \end{aligned}
\end{equation}
where $\mathbf{f}_r^{(i)}$ denotes the firing-rate vector of layer $i$, the subtraction and ReLU are applied elementwise, and $\mathbf{1}^\top(\cdot)$ sums over all neurons in the layer, $\alpha_{\min}$ and $\alpha_{\max}$ define the desired firing-rate interval and $r$ controls the regularization strength.
In the axonal-delay model (\cref{fig:test_acc_reg}A), introducing firing-rate regularization leads to a slight reduction in test accuracy in the dense setting ($\eta = 0$) compared to the unregularized baseline (see \cref{tab:performance_models_comparison}), while noticeably reducing the number of synaptic operations (SOPs). Under strong delay sparsity ($\eta = 0.8$), the same regularization results in a larger drop in accuracy, indicating a compounding effect between activity constraints and limited temporal resources, even though SOPs continue to decrease.
Dendritic-delay models exhibit a similar qualitative trade-off between accuracy and SOPs as a function of the maximum allowed firing rate. However, the impact of regularization on accuracy is more pronounced than in the axonal case.
Understanding the mechanisms underlying this increased sensitivity remains an open question and is left for future work (see \cref{sec:limitations}).

\section{Limitations}
\label{sec:limitations}

When trained under the same approach as prior DCLS work \cite{Hammouamri24_dcls}, axonal and dendritic delay models achieve performance comparable to synaptic delay learning. This is an unexpected behavior, as synaptic delays are, in principle, more expressive, enabling even a single neuron to detect a spatio-temporal pattern. This raises the question of whether the benchmarks considered here sufficiently exploit such expressivity. More temporally demanding datasets, such as those suggested in \cite{Zajzon25_SymSeqBench, Chen25_nsa}, could be better suited to reveal differences between delay mechanisms.
\par
We leave several questions to future work, including the abrupt performance degradation observed for all delay types beyond a certain maximum delay, the consistently stronger performance of axonal compared to dendritic delays despite similar theoretical expressivity, and the increased sensitivity of dendritic delays to delay sparsification and firing-rate regularization.

\section{Conclusion}
We showed that learning axonal or dendritic delays enables spiking neural networks with standard leaky integrate-and-fire neurons to achieve performance comparable to existing synaptic delay learning and adaptive neuron approaches on the GSC and SSC benchmarks. By associating delays at the neuron level, our models substantially reduce buffering requirements and maintain competitive synaptic operation counts and spike activity. We further demonstrated that strong delay sparsity leads to only modest performance degradation, highlighting the robustness and efficiency of this approach. These results suggest that axonal and dendritic delays provide a practical and resource-efficient mechanism for temporal processing in spiking neural networks, particularly for event-driven and neuromorphic hardware.

\section*{Acknowledgment}
This project is funded by the German Federal Ministry of Research, Technology and Space (BMFTR) under the NEUROTEC-II program (Grants Nos. 16ME0398K and 16ME0399). We thank Yuankang Zhao for valuable discussions on the project, as well as Ismail Khalfaoui-Hassani and Timothée Masquelier for insightful discussions on implementing the different delay types.

\printbibliography

\end{document}